\documentclass[runningheads]{llncs}

\usepackage{eccv}

\usepackage{eccvabbrv}

\usepackage{graphicx}
\usepackage{booktabs}

\usepackage[ruled]{algorithm2e} 

\SetAlFnt{\small}
\SetAlCapFnt{\small}
\SetAlCapNameFnt{\small}
\SetAlCapHSkip{0pt}

\usepackage[normalem]{ulem}  

\usepackage{xr}

\usepackage{diagbox}

\usepackage{graphicx}
\usepackage{soul}
\usepackage{enumitem}
\usepackage{tabularx}

\useunder{\uline}{\ul}{}

\usepackage{xspace}

\usepackage[accsupp]{axessibility}  

\usepackage[breaklinks,colorlinks,citecolor=eccvblue]{hyperref}

\usepackage[capitalize]{cleveref}
\crefname{section}{Sec.}{Secs.}

\usepackage{orcidlink}

\begin{document}

\title{GroundUp: Rapid Sketch-Based 3D City Massing}

\titlerunning{GroundUp}

\author{Gizem Esra {\"U}nl{\"u}\textsuperscript{1} \hspace{.12cm} Mohamed Sayed\textsuperscript{2} \hspace{.12cm} Yulia Gryaditskaya\textsuperscript{3} \hspace{.12cm} Gabriel Brostow\textsuperscript{1} }
\institute{\textsuperscript{1}University College London \hspace{.12cm} 
\textsuperscript{2}Niantic \hspace{.12cm}
\textsuperscript{3}PAI and CVSSP, University of Surrey\\}

\authorrunning{{\"U}nl{\"u} et al.}

\newcommand{\gu}[1]{{\color{red}\bf [GU: #1]}}
\newcommand{\ms}[1]{{\color{purple}\bf [MS: #1]}}
\newcommand{\yg}[1]{{\color{blue}\bf [YG: #1]}}
\newcommand{\refactor}[1]{{\color{gray} [Refactor: #1]}}
\newcommand{\revise}[1]{{\color{teal} [Revise: #1]}}
\newcommand{\todo}[1]{{\color{magenta}\bf [TODO: #1]}}
\newcommand{\update}[1]{{\color{red}#1}}
\newcommand{\verify}[1]{{\color{red}\emph{Check if correct:} #1}}
\newcommand{\old}[1]{{}}
 \newcommand{\new}[1]{{\color{black}{#1}}}

\maketitle

\begin{abstract}

We propose \emph{GroundUp}, the first sketch-based ideation tool for 3D city \emph{massing} of urban areas. 
We focus on early-stage urban design, where sketching is a common tool and the design starts from balancing\old{ planned} building volumes (masses) and open spaces. 
With Human-Centered AI in mind, we aim to help architects quickly revise their ideas by easily switching between 2D sketches and 3D models, allowing for smoother iteration and sharing of ideas.
Inspired by feedback from architects and existing workflows, our 
system takes as a first input a user sketch of multiple buildings in a top-down view. 
The user then draws a perspective sketch of the envisioned site.
Our method is designed to exploit the complementarity of information in the two sketches and allows users to quickly preview and adjust the inferred 3D shapes.
Our model\old{, driving the proposed urban massing system,} has two main components. 
First, we propose a novel sketch-to-depth prediction network for perspective sketches \emph{that exploits top-down sketch shapes}.
\old{Second, we amalgamate the complimentary \emph{but sparse} 2D signals to condition a customarily trained latent diffusion model.}
\new{Second, we use depth cues derived from the perspective sketch as a condition to our diffusion model, which ultimately completes the geometry in a top-down view.}
\old{The diffusion model works in the domain of a heightfield,}
\new{Thus, our final 3D geometry is represented as a heightfield,}
allowing users to construct the city \emph{``from the ground up''}. 
\old{We will release the code, datasets, and interface.}
\new{The code, datasets, and interface are available at \href{http://visual.cs.ucl.ac.uk/pubs/groundup/index.html}{visual.cs.ucl.ac.uk/pubs/groundup}.}
\end{abstract}


\section{Introduction}
\label{sec:intro}
Urban design has a deep impact on people's lives, and it epitomizes the opportunities to bring Human-Centered AI for Computer Vision~\cite{HCAI2022} into iterative design~\cite{jacoby2016drawing}. 
The loop of drawing and discussing buildings, and specifically sketching the buildings' \emph{masses}, \ie coarse shapes, is the crucial first stage of urban planning~\cite{pitts2012parametric}. ``Architectural design begins with a massing study''~\cite{LeytonMichael2001AGTo}, where the term ``massing'' is used for this stage because it locks in the long-term balance between constructed mass versus open space. In pilot interviews, architects said that existing 3D software for urban modeling is too cumbersome for ideation and does not support beginners\old{, yet recent AI-based tools do not meet their needs}. 

\begin{figure}
    \includegraphics[width=\textwidth]{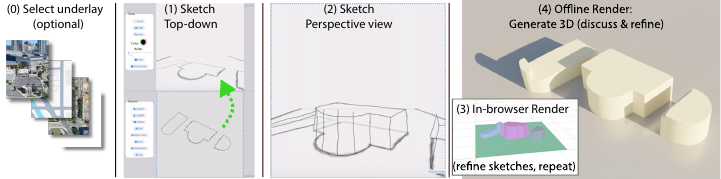}
    \caption{\textbf{An illustrative example of our method.} 
    (0) Users of our\old{ proposed} web-based GroundUp system can optionally load registered maps, satellite images, or perspective photographs as underlay layers. 
    These give context for the ``massing'' process.
    A blank underlay is used in this example. (1) (bottom) The user sketches the initial footprints of multiple buildings in a top-down view. These strokes are projected into the perspective-view canvas (top). (2) The user sketches a perspective view of the site, and then (3) they trigger our trained model to infer the 3D shape of the sketched buildings. \old{Finally (not shown), the user returns to sketching, refining their ideas, and repeatedly visualizing them in both 2D and 3D.} 
    \new{The user can then refine their ideas, iterating between 2D sketching and 3D visuals.}
    }
    \label{fig:teaser}
\end{figure}

Presently, the ease of sketching is hard to beat. 3D model precision is not the top priority in massing. Rather, urban design aims to satisfy the constraints and desires of whole teams of stakeholders. For example, an architect will play with many massing alternatives, often changing their mind mid-sketch. 
Currently, they iterate further in 2D with fellow architects on a shortlist of favorites, before re-doing just one or a few designs in 3D software (\eg Rhino or Sketchup), to test out the idea\old{, ideally with surrounding terrain and structures}. 
\old{Visualization of the 3D should not be a bottleneck for urban design.}
\new{Our work aims to facilitate the design process by providing the means to quickly preview designs in 3D.}

We propose GroundUp, a sketch-based 3D modeling tool for city massing. As shown in \cref{fig:teaser} and the video, it works by getting the user to draw and refine their ideas in two views: a top-down ``plan'' sketch and a perspective sketch. In both views, users can optionally sketch on top of backprojected lines and selected underlay photos. 
\old{This helps iterate, and for example, when remodeling an existing site.}
\new{This helps to iterate or when remodeling of an existing site is required.}
Tightly coupled with this interface, our algorithm quickly infers 3D massing-quality geometry. 
Such 3D geometry, once approved, can be refined outside of GroundUp and used in the downstream stages of architectural design.  

The model intertwined with this interface faces multiple challenges. Compared to photos, sketch lines only provide a sparse signal about the scene. 
Between a top-down and perspective sketch, it is hard to expect texture regions to match in appearance, making off-the-shelf approaches targeting 3D reconstruction from multi-view images \cite{duzceker2021deepvideomvs, lipson2021raft, sayed22eccv} inapplicable to our problem.
Additionally, urban areas are inherently complex scenes, so perspective views that convey building heights and roof shapes also suffer from extreme self-occlusions. With many unobserved or partially-observed regions, we turn to diffusion as a generative formulation that could help our method to reconstruct plausible building shapes (\cref{fig:teaser}(3)).

Critically, the model updates must be responsive for the system to be usable, 
imposing\old{ some} trade-offs between interactivity and the geometric quality from our adapted latent diffusion model.

Our proposed solution to these challenges offers the following contributions:
\begin{itemize}[noitemsep,topsep=0pt]
    \item GroundUp is the first system for quick 2D sketch-based iteration on 3D massing design of city blocks. 
    \item Our novel sketch-to-depth prediction network for perspective sketches exploits the top-down view's cues, and necessitated a bespoke training data process for this important domain.  
    \item We carefully design our top-down diffusion model to handle multiple conditions, integrating cues from both top-down and perspective sketches.    
\end{itemize}


\section{Related Work}

Sketch-based 3D shape modeling systems greatly facilitate the creation of 3D content, and the first proposed systems came in the 1970s~\cite{clowes1971seeing,huffman1971impossible}.
For an in-depth review of existing systems, we refer the reader to the comprehensive reviews~\cite{bonnici2019sketch, bhattacharjee2020survey, camba2022sketch}. 
Here, we focus on works related to our overall goal of urban reconstruction and papers most related to our method.

\subsection{3D Building and City Reconstruction}

The exciting works related to the problems of urban reconstruction can be classified into two sets of problems\cite{feng2021review}: first of layout generation~\cite{benes2021urban,He_2023_ICCV,deng2023citygen} and second, of city modeling and rendering~\cite{mitra2017sigga, mitra2018sigga,kim2020citycraft,xie2023citydreamer,Lin_2023_ICCV }.
In our work, we aim to provide the user with direct control of the layout \new{via sketching in a top-down view}, rather than generating it automatically.
\old{Then}\new{Next}, many algorithms \cite{wang2021cvprw,stucker2022isprs,li2022point2roof} for architectural modeling take as input point clouds or Digital Surface Models (DSMs) that contain building height information, obtained with LiDAR (Light Detection and Ranging) or photogrammetry~\cite{zhao2023review}. 
In contrast, we pursue a different goal of how to obtain buildings' height information from \old{a }sparse user-provided sketch\new{es}.

Multiple works utilize convolutional neural networks (CNNs) for monocular depth estimation from a satellite image~\cite{ghamisi2018img2dsm, mou2018im2height, mahmud2020boundary, chen2023heightformer, chen2023htc} and building segmentation in a satellite image \cite{mahdi2020aerial,li2021geometry,chen2023large}, or both~\cite{mahdi2020aerial, li2021geometry}. 
In the first stage of our method, we also rely on a CNN to obtain a segmentation of a top-down sketch into individual buildings. 
We then propose to inject this information into a monocular depth estimation network that takes a perspective sketch as an input -- the step that we show is paramount in the context of sparse sketch inputs.


\subsection{3D from Sketches}

\paragraph{3D representations:}
Sketch to 3D inference has been based on voxel-based representations \cite{Delanoy2018cgit}, point clouds~\cite{Yue20203dv, Wang2022eccvWorkshop}, implicit functions \cite{zhong2022study, guillard2021sketch2mesh, chowdhury2022garment}, and 3D diffusion models~\cite{binninger2023sens}.
Existing methods have a restricted ability to reconstruct details and to scale to larger scenes (\eg multiple objects).
We aim for the prompt reconstruction of multiple object shapes within an interactive interface.  
Our method controls for computational complexity and reduces memory footprint by regressing only 2.5D information, which is subsequently converted to a 3D mesh.
We leverage depth and normal maps as intermediate representations.
Using intermediate representations such as depth and normal maps is a common approach in sketch-based 3D reconstruction \cite{su2018interactive, wu2018eccv, gao2022eccv}. 
Just as we leverage a U-Net architecture~\cite{ronneberger2015u}, several works do so to predict multi-view depth and normal maps \cite{lun20173dv,zhong2020towards,Li2018robust}.
These methods then fuse the maps to a 3D shape. 
\old{A}\new{In contrast, a}iming at complex scenes with multiple occlusions, we predict only one perspective view map and rely on a diffusion model to predict a plausible heightfiled, matching perspective and top-down views.  
Recent work targeting lifting sketches of machine-made shapes to 3D \cite{puhachov2023reconstruction}, similar to us, \old{predicts depth as the first step}\new{first predicts depth}.
Their full method focuses on the reconstruction of sharp edges. However, it takes about two minutes for single object inference on average. 
In comparison, our method runs end-to-end in under 2.7 seconds on multi-building scenes. 

\paragraph{Ambiguity of 3D reconstruction:} 
For single-view and even sparse multi-view reconstruction, \emph{unobserved} regions create uncertainty, on top of the shape ambiguity of the observed geometric surfaces. 
Learning of shape category priors is one of the most prominent approaches to dealing with sparse sketch inputs~\cite{Yue20203dv, zhang2021sketch2model, guillard2021sketch2mesh, Wang2022eccvWorkshop, 
cheng2022cross, binninger2023sens, LAS23}.
To further alleviate uncertainty, in single object modeling, symmetry can be leveraged \cite{yao2020front2back,hahnlein2022symmetry}.
 Other approaches regress parameters of predefined procedural programs~\cite{nishida2016interactive,pearl2022geocode}, or assume availability of additional information \cite{li2022free2cad, zheng2022deep}. 
For our task of modeling 3D building shape masses in city neighborhoods, we have more pronounced uncertainty from the multiple layers of occlusions in any perspective view. 
Additionally, 3D buildings and their groups also have irregular shapes covering diverse geometric configurations.
Therefore, we leverage a generative diffusion model that allows us to obtain plausible-looking building \emph{masses} from two sparse sketch constraints (\cref{fig:teaser}).

\paragraph{Urban modeling:}
Nishida~\etal~\cite{nishida2016interactive} explored data-driven inference of procedural grammars for individual building reconstruction. The reconstruction ability of such methods is limited to what is possible to represent with the considered grammar.
\old{Moreover, such an}\new{Also, their} approach cannot infer the shape from a complete drawing and assumes a specific drawing order, matching the grammar used.
Liu~\etal~\cite{liu2021buildingsketch} extends procedural modeling to VR sketch inputs.
Vitruvio~\cite{tono2022arxiv} targets individual 3D building reconstruction from input sketches.
The paper stresses the importance of \emph{perspective 2D sketch-based modeling in architectural applications for early idea development}. 
Their method adopts an occupancy network \cite{mescheder2019cvpr} that is either fine-tuned or trained from scratch on synthetic sketches of individual buildings. 
However, their results show blobby reconstructions with some floating geometry pieces, typical in implicit 3D shape representations such as occupancy grids and signed distance fields \cite{park2019deepsdf, luo20233d,nam20223d}. 
Our heightfield representation allows us to obtain higher quality reconstructions of multiple buildings in one scene.

\subsection{Depth Estimation from RGB Images}
Sketches are harder for shape inference than RGB images, 
but we draw lessons nonetheless. For calibrated stereo pairs~\cite{collins1996space, vzbontar2016stereo,kang2001handling} or unstructured views with known poses~\cite{goesele2006multi, furukawa2015multi, schonberger2016pixelwise}, cost volumes reveal metric depth by matching photometric appearance between views. 
Unfortunately, the winning disparities are misleading with our textureless sketches. 
For depth from a single image, recent methods rely on a learned prior for depth estimation~\cite{eigen2014neurips, zhao2020monocular, godard2017unsupervised}. Follow-ups utilize 3D point networks~\cite{yin2021learning} to combat scale ambiguity, dataset mixing~\cite{ranftl2020towards} for more generalizable models, classification heads~\cite{fu2018deep} for improved accuracy, or generative models~\cite{ke2023repurposing, Duan2023corr, Saxena2023corr} for sharper depth maps. 
Recent methods combine the two: cost volumes and strong image priors, to produce sharp metric depths from multiple views~\cite{duzceker2021deepvideomvs, lipson2021raft, sayed22eccv}. 
\old{Rather than relying solely on photometric matching, we utilize a \old{privileged }top-down sketch -- with a known \new{via our sketching interface} pose, relative to the perspective views -- in an occupancy volume to resolve scale ambiguity.}
\new{Rather than relying on photometric matching, we utilize a top-down sketch in an occupancy volume to resolve scale ambiguity.}


\section{Method}

\begin{figure*}[h] 
  \includegraphics[width=\textwidth]{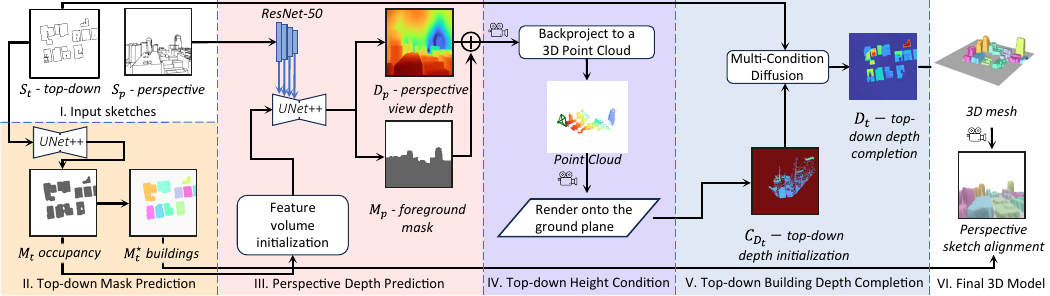}
  \caption{\textbf{Reconstruction pipeline overview.} 
  (I.) From input sketches, (II.) we estimate the segmentation of the top-down sketch into individual buildings (as detailed in \cref{sec:topdown_mask}). 
    (III.) We then inject the volumetric information about the spaces not occupied by buildings (based on the segmentation result and \new{using a known perspective camera from our interface}) into the network that predicts depth and a foreground mask for the perspective sketch view (further detailed in \cref{sec:perspective_depth}). 
(IV.) From the predicted depth values, we obtain a partial 3D point cloud of the user-envisioned 3D city block. 
(V)  By projecting a sparse 3D prediction into a top-down view, we obtain an initial guess for a top-down view \new{heightfield}.  Finally, we rely on a diffusion model to obtain a plausible 3D reconstruction that aligns with the perspective and top-down sketches (as shown in V-VI. and detailed in \cref{sec:diffusion_completion}).}
  \label{fig:pipeline}
\end{figure*}

Our supervised model is tightly coupled with the user-facing 2D and 3D interface described in \cref{sec:intro} and in \cref{fig:teaser}. 
The model has several components\old{.}\new{, shown and summarised in \cref{fig:pipeline}}. 

\subsection{Building occupancy mask estimation for top-down sketches}
\label{sec:topdown_mask}
First, given a top-down sketch $S_t$ we aim to obtain building occupancy $M_t$, and instance segmentation $M^{\star}_t$ maps (\cref{fig:pipeline} II.):
We use a subscript $t$ to denote maps of the top-down views.
Our top-down occupancy prediction network follows a UNet++ architecture: an encoder-decoder network with dense and nested skip connections introduced in \cite{zhou2018unet++}.
As an encoder, we use ResNet-50 \cite{he2016deep}, initialized with the weights of the model pre-trained on ImageNet~\cite{deng2009imagenet}.
We train our network with a weighted binary-cross entropy (BCE) loss
\begin{equation}
    \mathcal{L}_{mask} = - \frac{1}{N} \sum_{i=1}^{N} \Bigl[ \lambda_1 \left[ y_i \log(p_i) \right] + \lambda_0 \left[ (1-y_i) \log(1-p_i) \right] \Bigr],
    \label{eq:mask_bce}
\end{equation} 
where $y_i$ and $p_i$ are the ground-truth and predicted mask values for the $i^{th}$ pixel, respectively.
$\lambda_0$ and $\lambda_1$ are the weights for ground and building class predictions, respectively.
We empirically found that a bigger weight $\lambda_1$ for the building pixels improves mask prediction performance, accounting for class imbalance as buildings occupy a smaller area in the image.
We provide further implementation details in the supplemental.

We then segment into individual buildings, $M^{\star}_t$, by applying Connected-Component Labeling \cite{rosenfeld1966jacm} (\cref{fig:pipeline}~II.). 
We use this building-level segmentation $M^{\star}_t$ for visualization of the 3D reconstruction results \new{in our UI}\old{, as shown in }\new{ (}\cref{fig:pipeline}~VI\new{)}.

\subsection{Depth prediction from perspective sketches}\label{sec:perspective_depth}
Given a perspective sketch $S_p$ and a top-down building mask $M_t$, we aim to predict perspective depth maps $D_p$ and foreground masks $M_p$ (\cref{fig:pipeline}~III.) -- We use a subscript $p$ to denote maps of the perspective views.
In contrast to the masks in \cref{sec:topdown_mask}, $M_p$ labels both the building and ground pixels as foreground, with the background being sky pixels. 

\subsubsection{Network design} 
\label{sec:net_design_depth_p}
Predicting depth from a single sparse sketch is an ill-posed problem. 
Moreover, in our scenario, each sketch can be quite complex with multiple buildings and occlusions. 
We design our \new{perspective view }depth predictor to handle such complex urban scenes. 

Our architecture is inspired by a multi-view depth estimation method \cite{sayed22eccv}.
The backbone of this network is a UNet++ architecture, identical to the one we introduced in \cref{sec:topdown_mask}.
To reduce ambiguity in a perspective view, we leverage top-down view information. 
However, applying the multi-view stereo as in~\cite{sayed22eccv, yao2018mvsnet} is not feasible, as our views have little visual overlap so it is infeasible to perform meaningful feature matching between such views. 
Instead, we exploit the fact that the top-down building occupancy mask, $M_t$, provides information on whether a location in 3D space is free.
We construct a 3D occupancy volume, which is aligned with the perspective view frustum. 
We construct it by slicing the 3D view frustum of the perspective camera with $n$ depth planes at equidistant intervals between the near $d_\textrm{near}$ and far planes $d_\textrm{far}$. 
We populate the occupancy volume by setting all voxels that fall above non-occupied regions to $-\nu$ and all voxels above occupied regions to $\nu$. 
We discuss the choice of $\nu$ in detail in the supplemental material. 
Intuitively, we pick $\nu$ to be sufficiently large, but within the range of our encoder features. 
\new{We feed 3D occupancy features as input to the UNet++ encoder.}
\new{Namely, the 3D occupancy features are of shape $D \times H \times W$, where $D$ is the number of depth planes. 
When feeding these features into the 2D encoder in the UNet++, we consider depth planes as image feature channels $C$.}
Then, similarly to \cite{sayed22eccv}, we\old{ first} pass the input sketch through a ResNet-50 encoder to obtain multi-level features. 
\old{We then feed 3D occupancy features as input to the UNet++ encoder.} 
Starting from the first layer\new{ of the UNet++}, at every second layer, we concatenate output features with corresponding encoded sketch features. 
This network design allows us to efficiently inject top-down sketch information, resulting in more accurate perspective depth predictions.  
We provide the ablation study of our design in \cref{sec:ablation_depth_p}.

\subsubsection{Training}
During training, we use ground-truth $M_t$ building occupancy masks. 
We train our depth predictor with a weighted sum of four loss terms, so
\begin{equation}
   \mathcal{L}_{D} = \omega_d \textcolor{Cerulean}{\mathcal{L}_\textrm{depth}} + \omega_g \textcolor{Plum}{\mathcal{L}_\textrm{grad}} + \omega_n  \textcolor{ForestGreen}{\mathcal{L}_\textrm{p,norm}} + \omega_m  \textcolor{RubineRed}{\mathcal{L}_\textrm{mask}},
\end{equation}
where $\omega_{*}$ denotes the weight of the corresponding loss component. 
We introduce each term below.

$\textcolor{Cerulean}{\mathcal{L}_\textrm{depth}}$ is a multi-scale loss on depth predictions, that was shown to provide sharper depth maps at depth discontinuities than a loss applied only at the final depth map resolution  \cite{eigen2014neurips, godard2017cvpr, Godard2019iccv, sayed22eccv}. 
Following previous works, we predict depths at four resolutions from different levels of our UNet++ decoder, such that at each of the subsequent scales the spatial resolution is doubled.
It is defined as 
\begin{equation}
\textcolor{Cerulean}{\mathcal{L}_\textrm{depth}} = \sum_{s=1}^{S} \big\| (D_{p})_s - (D_{p}^{gt})_s \big\|_1
\end{equation}
where $\big\| \cdot \big\|_1$ is the $L_1$-norm and $(D_{ps}, D_{ps}^{gt})$ are the predicted and ground-truth depth maps at the $s^{th}$ scale. 

Similarly, inspired by \cite{li2018cvpr,sayed22eccv}, to encourage smoother gradient changes and sharper depth discontinuities in predicted depth maps, 
we use a multi-scale loss $\textcolor{Plum}{\mathcal{L}_\textrm{grad}}$ that penalizes differences in depth gradients between the predicted and ground-truth depth map:
\begin{equation}
\textcolor{Plum}{\mathcal{L}_\textrm{grad}} = \sum_{s=1}^{S} \big\| \nabla_x R_s \big\|_1 + \big\| \nabla_y R_s \big\|_1 ,
\end{equation}
where $R_s = (D_{p})_s - (D_{p}^{gt})_s$. 

Following Yin~\etal~\cite{yin2019iccv}, who showed that a geometric constraint on normal maps improves monocular depth estimation, we use a loss $\textcolor{ForestGreen}{\mathcal{L}_\textrm{p,norm}}$ between ground-truth $N_p^{gt}$ and predicted $N_p$ normal maps:
\begin{equation}
\textcolor{ForestGreen}{\mathcal{L}_\textrm{p,norm}} = \sum_{i=1}^{N} (1 -  (N_{p})_i \cdot (N_{p}^{gt})_i),
\label{eq:normal_loss}
\end{equation}
where we sum over the dot products of normal vectors $(N_{p}^{*})_i \in \mathbb{R}^3$ in corresponding normal map locations $i$.
We observed that this loss improves the performance in our setting as well. Both $N_p$ and $N_p^{gt}$ are computed on the fly from their corresponding depth maps.

Finally, $\textcolor{RubineRed}{\mathcal{L}_\textrm{mask}}$ is a weighted BCE loss, defined similarly to the one in \cref{eq:mask_bce}.
We use it to segment out building and ground pixels.

\subsection{Conditional \old{denoising }diffusion model for 3D building reconstruction}
\label{sec:diffusion_completion}

In the previous section, we described how we obtain a depth estimation for a perspective sketch view. 
As the next step, we backproject the depth map to obtain a 3D point cloud. 
From this point cloud, we initialize a heightfield of the city block aligned with the top-down user sketch. 
To account for possible inaccuracies in the depth prediction network, we leverage a mask $M_t$, predicted with our building occupancy mask estimation network as described in \cref{sec:topdown_mask}. 
We set all heightfield predictions that fall outside the occupied regions to a constant ground-level value.
We then use a diffusion model conditioned on the input sketch and the initial heightfield from the perspective sketch view to complete missing depth regions in the top-down view. Since our conditioning relies on both views, the model predicts plausible 3D buildings that align with user sketches.
Note that, during training, we initialize heightfields using ground-truth perspective view depth maps.

\subsubsection{Network architecture} 
\label{sec:design_diffusion}
We build on the latent space diffusion model by Duan~\etal~\cite{Duan2023corr}, adapting it to handle multiple conditions.
We chose a latent diffusion model due to its memory efficiency and inference speed compared to image-space diffusion models.

We map ground-truth depth maps to a latent space using a depth encoder: $z = \mathcal{E}_\textrm{depth}(D^{gt}_t)$. Additionally, we encode sketch and depth conditions: $c_\textrm{sketch}=\mathcal{E}_\textrm{sketch}(C_{St})$ and $c_\textrm{depth}=\mathcal{E}_\textrm{depth}(C_{Dt})$, respectively. 
We initialize $\mathcal{E}_\textrm{sketch}$ and $\mathcal{E}_\textrm{depth}$ with pre-trained weights. 
Specifically, for $\mathcal{E}_\textrm{sketch}$, we employ a ResNet-50 architecture pre-trained on ImageNet.
For the depth encoder, we employ the one from the Stable Diffusion \cite{rombach2022cvpr}. We pre-train the autoencoder following their strategy and supervise using ground-truth top-down depth maps $D^{gt}_t$ using KL-regularization in the latent space. We fine-tune both latent encoders when training the full model.

To construct the final input to the denoising network, we combine the sketch, $c_\textrm{sketch}$, and depth, $c_\textrm{depth}$, conditions with the noisy depth latent $z_{k}$, for a given noise level $k$. 
To align features, we pass latent representations $c_\textrm{depth}$ and $z_{k}$ through two separate CNNs, consisting of two convolutional layers.
The final denoising network input is created by combining sketch latent features and depth conditions with the noisy depth latent through an element-wise summation.

\subsubsection{Training}
The training objective for the diffusion process is defined as 
\begin{equation}
    \mathcal{L}_\textrm{diff} = \mathbb{E}_{k \sim [1,T], z_k, \epsilon_k} \left[ \| \epsilon_k - \epsilon_\theta (z_{k}, c_{S_t}, c_{D_t}, k) \| \right]^2,
    \label{eq:diffusion_loss}
\end{equation}
where $\epsilon_t$ and $\epsilon_\theta$ are the ground-truth and predicted noise maps, at timestep $k$.

Additionally, we use auxiliary pixel-based losses to help train the conditioning process. Firstly, $L_1$ and $L_2$ losses on predicted $D_t$ and ground-truth $D_{t}^{gt}$ depth maps are used, defined as 
\begin{equation}
    \mathcal{L}_{L_1} = \big\| D_{t} - D_{t}^{gt} \big\|_1 \quad \textrm{and} \quad   
    \mathcal{L}_{L_2} = \big\| D_{t} - D_{t}^{gt} \big\|_2.
    \label{eq:losses_l1_l2}
\end{equation}
We also use a loss on normal maps $\mathcal{L}_{t,norm}$, defined similarly to the one in \cref{eq:normal_loss}.
We find that this loss results in sharper, more uniform depth predictions.
We ablate its effect in \cref{sec:ablation_diffusion}.

The complete objective loss of our top-down heightfield completion diffusion model is defined as 
\begin{equation}
    \mathcal{L}_\textrm{total} = \mathcal{L}_\textrm{diff} + \mathcal{L}_{L_1} + \mathcal{L}_{L_2} + \mathcal{L}_\textrm{t,norm}.
    \label{eq:diff_loss_full}
\end{equation}

\subsubsection{3D mesh: From the ground up}
\label{sec:mesh_create}
Finally, to obtain a 3D mesh, we create a 3D mesh grid $\mathcal{M}^{3D} \in \mathbb{R}^{N \times N \times 3} $ with $N \times N$ vertices, where $N$ is the width/height of the top-down depth map where the horizontal $x$ and vertical $y$ axes map to pixel coordinates.
We obtain the height of each vertex $v_{ij}$ in $\mathcal{M}^{3D}$ as 
\begin{equation}
    v^{z}_{ij} = d_\textrm{ground} - (D_t)_{ij},
\end{equation}
where $d_\textrm{ground}$ is the depth value of the ground plane and $(D_t)_{ij}$ is the predicted top-down depth at pixel location $(i,j)$. 
We assign $d_\textrm{ground}$ to the maximum depth value in $D_t$.


\section{Experiments}
\label{sec:results_synth}

In this section, we evaluate our method on synthetic sketches. We first evaluate our perspective depth prediction network and discuss the importance of various design choices. 
We then assess our complete method, by evaluating our top-down completion network on inputs predicted by the perspective depth network. 
We compare with a few alternative baselines and ablate our design choices.
The details of data generation and splits are provided in the supplemental.

\subsubsection{Perspective depth prediction}
\label{sec:ablation_depth_p}

In \cref{table:sketchrecon_metrics}, we assess our design choices for the perspective depth prediction network and compare against several baselines using standard depth metrics~\cite{eigen2014neurips}. Briefly, \textit{Abs Diff} is the absolute difference between ground-truth and predicted depth maps, \textit{Abs Rel} is the absolute difference normalized by the ground-truth depth map, \textit{Sq Rel} is the square of \textit{Abs Rel}, \textit{RMSE} is the root mean square error between both depth maps, \textit{Log RMSE} is the root mean square error on logged depths, and \textit{a5} is the ratio of pixels whose depth values have a relative depth error lower than 5\%.

\begin{table}[t]
\centering
\begin{tabular}{l|l|rrrrrr}
\multicolumn{1}{c|}{\textbf{}} & \multicolumn{1}{c|}{\textbf{Model}} & \multicolumn{1}{c}{\textbf{\begin{tabular}[c]{@{}c@{}}Abs Diff $\downarrow$ \end{tabular}}} & \multicolumn{1}{c}{\textbf{\begin{tabular}[c]{@{}c@{}}Abs Rel$\downarrow$ \end{tabular}}} & \multicolumn{1}{c}{\textbf{\begin{tabular}[c]{@{}c@{}}Sq Rel$\downarrow$\end{tabular}}} & \multicolumn{1}{c}{\textbf{RMSE$\downarrow$}} & \multicolumn{1}{c}{\textbf{\begin{tabular}[c]{@{}c@{}}Log RMSE$\downarrow$\end{tabular}}} & \multicolumn{1}{c}{\textbf{ a5$\uparrow$}} \\ \hline
1 & $Mono_S$ & 6.64 & 5.33 & 0.79 & 10.06 & 9.41 & 66.13 \\
2 & $Mono_L$ & 5.57 & 4.33 & 0.53 & 8.58 & 7.44 & 68.82 \\ \hline
3 & $OV_{S}$ & 6.23 & 5.24 & 0.74 & 9.28 & 9.55 & 67.80 \\ \hline
4 & $OV_{L}$ & \textbf{3.49} & \textbf{2.13} & \textbf{0.21} & \textbf{6.54} & \textbf{3.43} & \textbf{89.20} \\ \hline
\end{tabular}%
\caption{\textbf{Quantitative evaluation of the perspective depth estimation.} 
$Mono$ stands for a monocular depth predictor baseline by Sayed \etal~\cite{sayed22eccv}, where subscripts $S$ and $L$ define a smaller and larger encoder backbones, respectively.
$OV$ represents our model with the occupancy volume obtained as described in \cref{sec:net_design_depth_p}.
Please see \cref{sec:ablation_depth_p} for the details.
All metric values apart from $a5$ are scaled up by $10^2$.
}
\label{table:sketchrecon_metrics}
\end{table}

\paragraph{Baselines:} 
We train a naive monocular depth predictor baseline from Sayed~\etal~\cite{sayed22eccv} without a cost volume (no source views for multi-view stereo), which we refer to as $Mono$, and compare two image encoder backbones. 
In \cref{table:sketchrecon_metrics}, lines [1-2] refer to $Mono_S$ for a smaller (EfficientNet~\cite{tan2019efficientnet}) and $Mono_L$ for a larger encoder (ResNet-50 \cite{he2016deep}). 
A large image encoder leads to superiority across all depth metrics, with a minimal increase in inference speed -- \textit{~0.16}s on average per sample. Given this, we use this larger backbone for all other experiments.

\paragraph{Ablations:}
In \cref{table:sketchrecon_metrics}, $OV$ represents our model with the occupancy volume obtained as described in \cref{sec:net_design_depth_p}.
\new{We empirically found $\nu=50$ to give the best results. This value is close to the mid-point of the range of multi-scale image features. We hypothesize that this setting allows the network to leverage the occupancy information most beneficially.}
We provide a detailed analysis of the choice of $\nu$ in the supplemental.
\old{Here we compare two settings: $\nu=1$ and $\nu=50$, where the latter is aligned better with the range of image features from the ResNet-50 encoder.}  
\old{Lines [6] vs.~[5] show that $\nu=50$ is better for the larger encoder backbone, which is ResNet-50 in our case.}
\new{Lines [4] vs.~[3] show the advantage of the larger encoder backbone.} 
Our complete model then comprises a ResNet-50 encoder backbone, and an occupancy volume with voxels assigned using $\nu=50$. \old{\cref{table:sketchrecon_metrics} shows the superiority of this model over alternative baselines.}

\begin{figure}[h]
  \includegraphics[width=\textwidth]{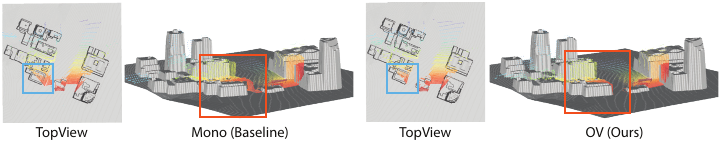}
  \caption{\textbf{Qualitative evaluation of the perspective depth estimation.} 
$Mono$ stands for a monocular depth predictor baseline by Sayed \etal~\cite{sayed22eccv}. 
$OV$ represents our model with the occupancy volume, obtained as described in \cref{sec:net_design_depth_p}. 
Grey mesh corresponds to the geometry obtained from the ground-truth heightfield. 
Point clouds represent the estimated depth values from a perspective sketch.
Colors encode the distance from a camera.
Our prediction visually aligns better with the ground-truth.
  }
  \label{fig:depth_mono_vs_ov}
\end{figure}

\paragraph{\new{Comparison:}} 
In \cref{table:sketchrecon_metrics}, lines [3-4] vs [1-2] show that the $OV$ models outperform $Mono$ models. 
We show a qualitative comparison \new{of the $Mono$ baseline with our $OV$ method} in \cref{fig:depth_mono_vs_ov}, showing the importance of the proposed occupancy feature volume for correcting for spatial ambiguity from single-view depth estimation. 

\begin{table}[t]
\centering
\begin{tabular}{lccccc}
\cline{1-5}
\multicolumn{1}{l}{ \textbf{Model}}             & \textbf{Abs Diff↓} & \textbf{Abs Rel↓} & \textbf{Sq Rel↓} & \textbf{RMSE↓}  \\ 
\cline{1-5}
 $S_t$+$C_{Dt}$ (fs)                               & 0.1090          & 0.0221          & 0.0042          & 0.1246          \\
 $S_t$+$C_{Dt}$ (pt)                               & 0.1079	        & 0.0218	      &0.0040	        & 0.1221          \\
 $S_t$+$C_{Dt}$ (pt) + $\mathcal{L}_\textrm{t,norm}$ & \textbf{0.1056} & \textbf{0.0214} & \textbf{0.0039} & \textbf{0.1200} \\
 \cline{1-5}
 \textit{HeightFields}~\cite{watson2023wacv}   & 0.1099          & 0.0225          & 0.0046          & 0.1341          \\
\cline{1-5}
\end{tabular}
\caption{\textbf{Quantitative analysis of top-down depth prediction.} (fs) denotes training sketch and depth encoders from scratch jointly with the diffusion model. 
(pt) refers to pre-trained \old{latent }encoders for sketch and depth conditions, as described in \cref{sec:design_diffusion}.
 $S_t$+$C_{Dt}$ denotes that we use two conditions: a top-down sketch and a partial top-down depth prediction based on the perspective sketch view. 
 The numbers in the first two lines represent diffusion models trained with the losses defined by equations \cref{eq:diffusion_loss,eq:losses_l1_l2}, while the last line represents the model trained with the full loss \cref{eq:diff_loss_full}.
}
\label{table:diff_metrics}
\end{table}

\begin{table}[t]
\centering
\begin{tabular}{lccccccc}
\cline{1-7}
\textbf{Model} & \textbf{Acc↓} & \textbf{Compl↓} & \textbf{Chamfer↓} & \textbf{Precision↑} & \textbf{Recall↑} & \textbf{F Score↑}  \\
\cline{1-7}

 $S_t$+$C_{Dt}$ (fs)                               & 3.89            & 2.91            & 3.40            & 81.6          & 87.7          & 84.2          \\
 $S_t$+$C_{Dt}$ (pt)                               & \textbf{3.79}           & \textbf{2.90}           & \textbf{3.34}            & \textbf{81.8} & \textbf{88.1} & \textbf{84.4} \\
 $S_t$+$C_{Dt}$ (pt) + $\mathcal{L}_\textrm{t,norm}$ & 4.35            & 3.54            & 3.94            & 76.1          & 82.0          & 78.3          \\
\cline{1-7}
 \textit{HeightFields}\cite{watson2023wacv}   & 5.35            & 6.99            & 6.17            & 70.4          & 63.9          & 66.0          \\
\cline{1-7}
\end{tabular}
\caption{\textbf{Quantitative 3D evaluation of the final reconstructed meshes.} The metrics in this table account for the visibility of 3D geometry in a perspective sketch view\old{: We can only make a reliable guess on the heights of the buildings observed in these sketches}. Please see \cref{sec:ablation_diffusion} for details. The notation in this table matches the caption of \cref{table:diff_metrics}. 
We details on the metrics: \textit{Completion}, \textit{Accuracy}, \textit{Chamfer Distance}, \textit{Precision}, \textit{Recall}, and \textit{F-Score} can be found in~\cite{bozic2021transformerfusion}. 
\old{\revise{Although adding a normal loss leads to better depths (see \cref{sec:ablation_diffusion}) and makes reconstructions visually sharper, qualitative numbers are worse. This is likely due to shrunk vertical walls (a large number of evaluated points) in areas where sharp depth map discontinuities are encouraged.}}
}
\label{table:diff_3D_metrics}
\end{table}

\subsubsection{Top-down depth completion}
\label{sec:ablation_diffusion}

\begin{figure*}[t]
\includegraphics[width=\linewidth]{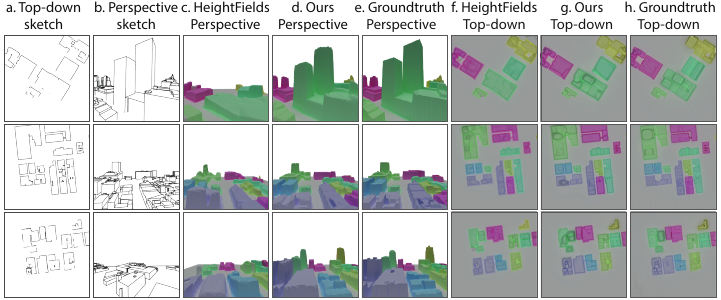}
  \caption{\textbf{Qualitative evaluation on synthetic sketches.} (a) and (b) show example top-down and perspective sketches. (c) and (f) show reconstruction results obtained with the \emph{HeightFields}~\cite{watson2023wacv} method, which is trained and tested on the same data as our method. (d) and (g) show reconstruction results by our method. 
  \old{Finally, }(e) and (h) show the heightfield of the ground-truth top-down depth map.
  \new{Note that the colors are assigned according to the ground-truth segmentation of buildings. Please zoom in to better see the alignment of predicted geometries with the ground-truth buildings' areas.}
  }  
  \label{fig:qualitative_diffusion}
\end{figure*}

Our final goal is to infer plausible building geometries from top-down $S_t$ and perspective $S_p$ sketches (\cref{fig:qualitative_diffusion} [a,b]).
Namely, we rely on the top-down sketch to recover building layouts and on the perspective sketch to estimate buildings' heights. 
We obtain height cues with the perspective depth prediction network. 
Then, the aim of our diffusion model, introduced in \cref{sec:diffusion_completion}, is to produce top-down depth maps faithful to the top-down sketch $S_t$ and height cues $C_{Dt}$.

\new{We first ablate the design of our network and then compare it with a deterministic baseline. For evaluations, we use metrics in 2D (\cref{table:diff_metrics}) and 3D (\cref{table:diff_3D_metrics}), comparing against the ground-truth.} 
\new{For 2D evaluation, we use metrics similar to the ones in \cref{table:sketchrecon_metrics}. 
Since we focus on buildings and not the terrain, we compute all 2D metrics only within buildings' ground-truth regions, using building masks $M_t$.}
\new{We evaluate 3D metrics only for the parts of geometries observed in the perspective sketch viewpoints.}
\new{This allows us to focus the evaluation on regions for which the perspective sketches provide explicit control of the buildings' heights.}
\old{Since top-down and perspective sketches provide only partial cues on the heights of the buildings, we evaluate 3D metrics only for the parts of geometries observed in the perspective sketch viewpoints.}
Before computing sampled point cloud distances between predicted and ground-truth meshes, we remove points not in the region around the back-projected ground-truth perspective depth map.

\paragraph{Role of pre-training:} 
\cref{table:diff_metrics} demonstrates the importance of pretraining sketch and depth encoders, $\mathcal{E}_\textrm{sketch}$ and $\mathcal{E}_\textrm{depth}$, respectively.
(fs) refers to training the encoders from scratch jointly with the diffusion model and (pt) refers to pre-training latent encoders for sketch and depth conditions.

\begin{figure}[t]
  \centering
    \includegraphics[width=\linewidth]{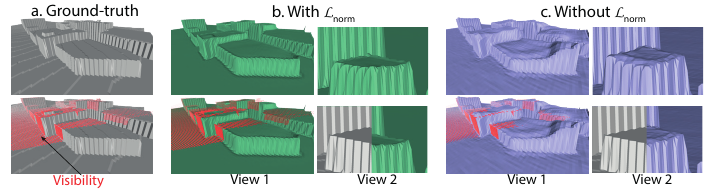}
   \caption{\textbf{Role of the normal loss.} \new{a) Visibility regions (red points) are computed based on ground-truth geometry and the perspective sketch viewpoint. b) Prediction when the normal loss is used: The red point cloud is riding slightly above the green prediction. As shown in \emph{view 2} the height is slightly underestimated in the visible regions, but the loss results in more even roofs overall. 
   c) Prediction when the normal loss is not used: the model produces blobby building geometry outside the visible regions.}} 
   \label{fig:normal-loss-m}
\end{figure}

\paragraph{Role of normal loss:} 
\old{Additionally, w}\new{W}e show the qualitative evaluation of the role of the normal loss in \cref{fig:normal-loss-m}.
It shows that the normal loss yields building geometries with sharper corners and flat building tops. 
Computed on all building regions, 2D losses in \cref{table:diff_metrics} show that the normal loss $\mathcal{L}_\textrm{t,norm}$, defined with \cref{eq:diff_loss_full}, significantly improves the accuracy of top-down depth-predictions -- reflecting on the overall appearance of the buildings. 
3D metrics in \cref{table:diff_3D_metrics}, computed only on visible regions from the perspective sketch viewpoint, highlight a slight geometry shrinkage, visible in View 2 in \cref{fig:normal-loss-m}.
While adding a normal loss hurts quantitative 3D metrics, we advocate its usage as it produces much sharper \new{and smoother} surfaces, as shown in \cref{fig:normal-loss-m} and supported with \cref{table:diff_metrics}.

\paragraph{Comparison with a deterministic baseline:}
Qualitative results of our method are shown in \cref{fig:qualitative_diffusion} (d) and (g): 
We can infer realistic building geometries following input sketches that closely resemble the ground-truth -- \cref{fig:qualitative_diffusion} (e) and (h). 
We compare our generative approach against the \textit{HeightFields}~\cite{watson2023wacv} baseline -- a deterministic model designed for heightfield completion from multi-frame RGB sequences. 
We train and test it on the same input as our model and visualize the results in \cref{fig:qualitative_diffusion} (c) and (f). 
In particular, \new{the \textit{HeightFields} model's test time input is the output of our first step: the predicted partial point cloud from a perspective view. This model is not a suitable stand-alone method for the task.}
\old{t}\new{T}o train this model, we also added $\mathcal{L}_\textrm{t,norm}$, as we found it to result in better performance.
However, even with this additional loss, the \textit{HeightFields} model fails to produce buildings with correct heights, and produces less plausible building geometries.
In particular, it fails to capture sharp details and flat rooftops. 

\cref{table:diff_metrics,table:diff_3D_metrics} show quantitative comparison of our full model with \textit{HeightFields}~\cite{watson2023wacv} baseline\old{, using metric computed on top-down depth maps}.
They show the superiority of our diffusion model in all settings, confirming the visual observations.

\old{All our models outperform the \textit{HeightFields}~\cite{watson2023wacv} baseline.}



\section{User Study}
\label{sec:results_users}
\paragraph{Modeling Interface:}
To validate our contributions, we built an interactive user interface in HTML, JavaScript, and Python. The 3D massing system runs real-time on a Titan X and can be used on any touch-screen device thorough a browser, ideally with a stylus. Broadly, the UI lets users sketch perspective and top-down views on 2D canvases, edit strokes, project a top-down sketch into the perspective canvas to align their sketches, and contains a 3D viewer. An overview of the UI is in \cref{fig:teaser} and is described in greater detail in the supplemental.

\paragraph{Evaluation:} To validate our system, we run a proof-of-concept user study. For the study, we collaborated with one of the world-leading schools in urban design\new{, the Bartlett School of Architecture, at University College London}. We engaged 5 urban design architects: 2 undergraduate students, advanced in their studies, and 3 postgraduates with varying years of professional practice. 
Additionally, to test how friendly our system is for users with limited modeling and sketching experience, we engaged 5 further volunteers. All users watched a short video tutorial and had 5 minutes to play with the interface before starting the task. 
To have a concrete qualitative goal in our main study, we chose to provide participants with reference top-down and perspective renderings as underlays (the example screenshot is provided in the supplemental).
We selected 9 scenes, randomly distributed between participants. Each participant drew two scenes.

In a post-study questionnaire, all architects indicated that they were able to recreate the building from the reference in under 5 minutes. 
As expected, it was more challenging for novices, yet, 2/5 were satisfied with the outcome.
On a 5-point Likert scale, architects (novices) gave an average score of 0.8 (1.2) on how well the results match the reference, with $+2$ for matching the reference well and $-2$ for failing completely. 
On a 5-point Likert scale, architects gave an average 1.4 score on how likely they are to use such an interface: where $-2$ for highly unlikely and $+2$ for highly likely.
This analysis shows that overall our system achieves a set goal of fast prototyping of building masses, while future work could aim to further improve the reconstruction accuracy.
The detailed statistics for the post-study questionnaire are provided in the supplementary.

\old{We considered but rejected doing a head-to-head comparison against Rhino, as i}\new{I}n our pilot study, architects consistently indicated that it would take them about 10 minutes in Rhino for scenes comparable to the ones we target. 
\new{The pilot study on SketchUp, documented in the supplemental, similarly showed that it is not suitable for fast prototyping.}
This, in particular, shows the lack of convenient tools for early design stages and reinforces the motivation for our work.

\begin{figure}[t]
  \includegraphics[width=\linewidth]{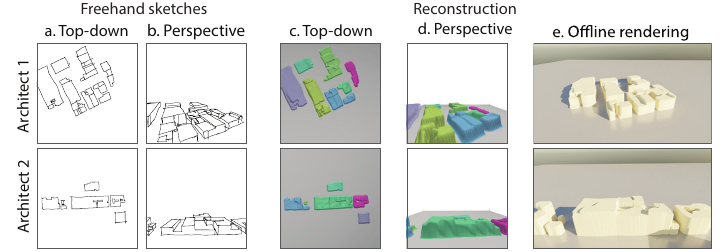}
  \caption{Freehand sketches and corresponding 3D reconstructions in our user interface, made by urban design architects in our study. (e) shows automatically post-processed results, rendered with an offline rendered, as described in the supplemental.
  }
  \label{fig:user_study_reference}
\end{figure}

Two urban design architects also did \emph{freehand modeling} after the main study and completed post-study questionnaires. These sketches are shown in \cref{fig:user_study_reference}.

\section{Conclusion \new{and discussion}}
\label{sec:conclusion}
We have presented the first sketch-based method for early-stage urban design, aligning it with the Human-Centered AI philosophy~\cite{shneiderman2022human}.
Taking into account design workflows that commonly start from top-down city layouts, we proposed models that, while working in image space, efficiently leverage information from both perspective and top-down sketch views. GroundUp addresses the especially challenging (but not unique) aspects of our problem: complexity and diversity of scene geometries, sparsity of sketch inputs, and incomplete depth cues in user-provided views.
\old{We evaluated our method numerically on synthetic data and qualitatively via a user study, leveraging our proposed interactive interface.} 
\new{While we only show the results for a single perspective sketch, our system is trivially extended to a multi-view setting: 
by projecting point clouds inferred from extra perspective sketches into the top-down views passed to our diffusion model. 
We provide numerical experiments in the supplemental.
With this work, we have taken a step toward quick building massing. To propel the integration of our tool into design workflows, future work might focus on directly predicting editable mesh representations and supporting finer details.
Additionally, it could be interesting to extended this work to trees and terrain, for example, by sketching trunks and contour lines for the terrain.
}

\new{\section*{Acknowledgements} 
We thank Prof.~Tobias Ritschel for his invaluable feedback and help; Natalia Laskovaya for an inspiring and detailed early discussion on design processes in architecture; Sharon Betts for her huge help in making our user study possible. 
We also thank Kening Guo and all the anonymous participants of the user studies.  
Gizem Esra {\"U}nl{\"u} is funded by a Niantic PhD scholarship.}

%
%
\bibliographystyle{splncs04}
\bibliography{MAIN-reference}

\section{Synthetic data}

\subsection{Data generation}
\label{sec:data-gen}
To train our networks, we use the UrbanScene3D dataset \cite{lin2022eccv} which contains large-scale 3D models of six real-world cities. 
We selected New York and Chicago for training and validation respectively, and San Francisco for testing. 
Our training set contains $40K$ samples, our validation set contains $2K$ samples, and our test set comprises $1K$ samples. 
For training samples, we perform random augmentation on the heights of individual buildings by scaling each building along the vertical axis to increase the diversity of our scenes. 
We generate synthetic sketches of buildings in perspective views and their ground-truth depth and segmentation maps, using Blender Freestyle~\cite{Blender}.

\subsection{View Selection in 3D cities} 
\label{sec:view_selection}
As we mentioned in \cref{sec:data-gen} of the main paper, we generate synthetic sketches of buildings in perspective views and their ground-truth depth and segmentation maps, using Blender Freestyle \cite{Blender}. 
We place two cameras for each scene: one top-down orthographic $Cam_t$ and one aerial perspective $Cam_p$. 
We start by sampling $Cam_p$'s location in the scene and consequently set $Cam_t$ in the positive look-at direction of the former at the midpoint between near and far planes. 
To avoid placing $Cam_p$ within building geometry, we pre-process each city and label traversable regions on the ground plane. 
Moreover, each camera sits at a pre-determined height above the ground, selected so that most of the buildings are observed from above. 
This implies that since our system was trained with fixed settings for top-down $S_t$ and perspective $S_p$ sketches, it expects that the inputs should adhere to these rendering settings. 
We recognize that this limits the choice of viewpoints, and in full-featured applications, the urban designer may want to choose a different viewpoint, such as a street-level sketch, or use an axonometric projection.
However, we believe that robustness to such representation changes is only a matter of training on a dataset that includes a wider range of rendering settings.

\begin{figure}[]
  \centering
  \includegraphics[width=0.92\linewidth]{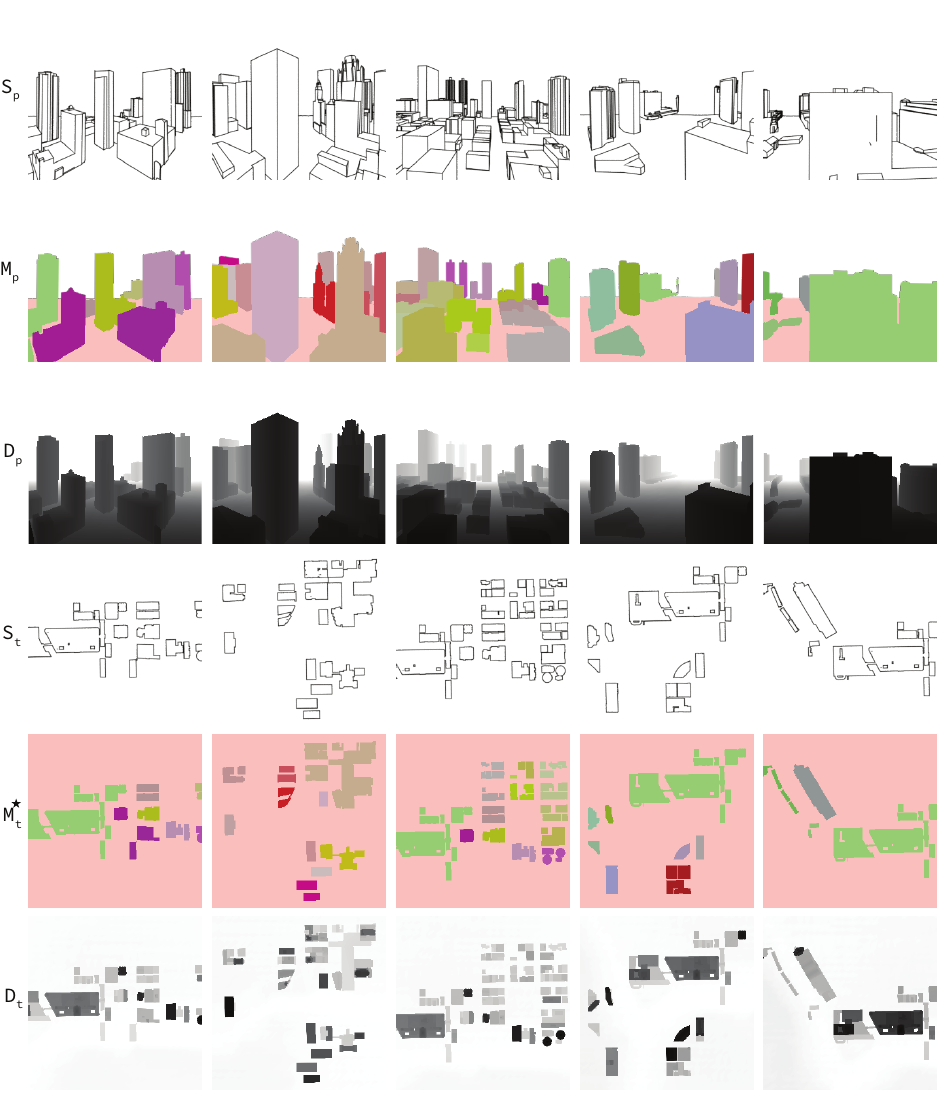}
  \caption{Ground-truth maps from our synthetic dataset: perspective synthetic sketches $S_p$, foreground masks for perspective views with visualized segmentation of buildings (in the method we only use the binary foreground mask), depth in perspective views $D_p$, top-down synthetic sketches $S_t$, building-level segmentation in top-down views $M^{\star}_t$ and top-down depth maps $D_t$.
  Please see \cref{sec:data-gen} of the main paper and \cref{sec:view_selection} for details on data generation and view selection.
  }
  \label{fig:dataset_samples}
\end{figure}

\subsection{Representative samples}
In \cref{fig:dataset_samples}, we provide samples from our training dataset, showing: perspective synthetic sketches $S_p$, foreground masks for perspective views $M_p$, depth in perspective views $D_p$, top-down synthetic sketches $S_t$, building-level segmentation in top-down views $M^{\star}_t$ and top-down depth maps $D_t$.

\begin{figure}[]
  \includegraphics[width=0.98\linewidth]{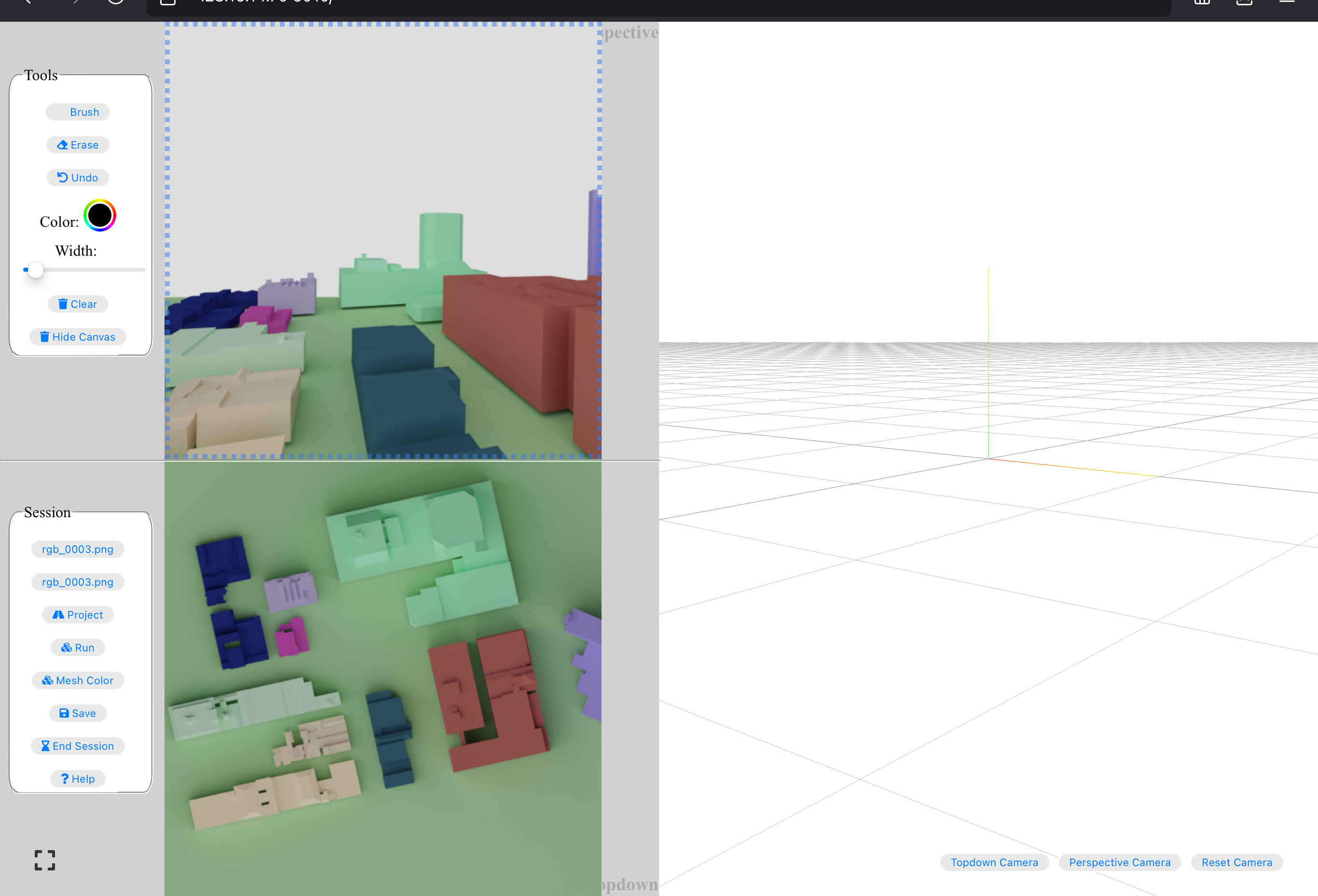}
  \caption{Screenshot of our interface, as seen by participants in our user study. Note that both sketching views (top-down and perspective) have been loaded with reference images. This means user study participants were mostly modeling existing massing, instead of inventing new designs.}
  \label{fig:UserStudyScreenshot}
\end{figure}

\section{User study: Modeling interface}

To validate our contributions, we built an interactive user interface in HTML, JavaScript, and Python. The 3D massing system works in real-time on a Titan X, and can be used on any touch-screen device through a browser. While mouse, touch, and stylus inputs are allowed, we recommend users use a stylus, because it is easier and results in higher-quality sketches. 

\begin{figure}[]
  \includegraphics[width=\textwidth]{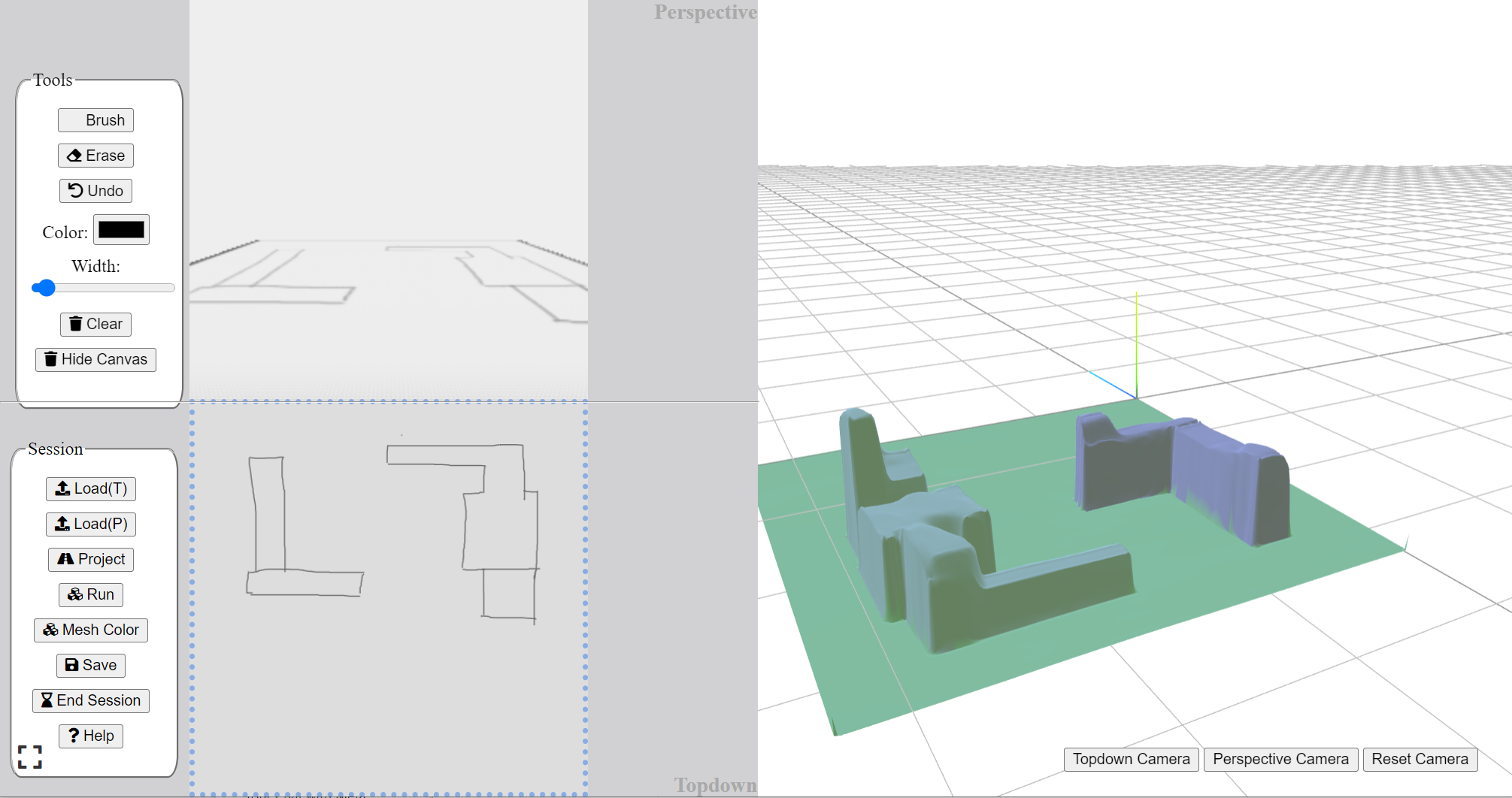}  
  \caption{Sketching interface. The interface is split into 3 major components: a 3D view for interaction with a predicted 3D scene, and two sketching canvases for perspective and top-down views. The buildings are generated here using only layout information.}
  \label{fig:interface_ui}
\end{figure}

The interface is split into three main regions: a 3D viewport for interaction with a predicted 3D scene, and two sketching canvases for perspective and top-down views.
\cref{fig:UserStudyScreenshot} and \cref{fig:interface_ui} show our sketching interface, with and without a reference underlay in the sketch canvases, respectively. 
For sketching, our tool supports standard capabilities such as erasing and undoing. Strokes are treated as vector data. 
We support two levels of zoom for the sketch views. 
The integrated 3D viewer is simple, and generated meshes can be exported to downstream 3D tools, \eg for adding details or vectorized rendering like \cref{fig:teaser} (4) of the main paper. 
An important capability that was added in response to pilot users was a sketch-to-sketch projection. Users can project their top-down sketches to the perspective canvas, allowing them to see the building layouts as a kind of foundation \cref{fig:interface_ui}. 
This supports users in aligning their masses between top-down and perspective sketches, which can be hard to do otherwise.

\section{User study: Additional details and feedback}

User feedback, especially from the post-study questionnaire, is provided in \cref{tab:UserStudyPostQuestionsSummary}. Additionally, we list here quotes from all the users in our study, grouped by sentiment: good, bad, and neutral. 

\subsection{Quotes with positive sentiment} 
\emph{
\begin{itemize}
    \item Very cool system!
    \item Make it easy to iterate on designs. I can adapt it as I go. Deterministic behavior, so I feel that I have control over the output.
    \item I think that could be a very useful tool. Even if I sketched it really really well on paper, I'll subconsciously convince myself it works, even if in 3D it doesn't work. (e.g. gaps in their Snake-tunnel model)
    \item For massing, we always start designing from the topdown and in sketch form. I usually have an idea in mind for what the design would look like in perspective. But just from the topdown it is hard to visualize how the buildings would look like in perspective. This tool is great for visualizing the perspective view quickly.
    \item This was FAST! In Rhino I'd need at least 10 min for basic meshforming, and then 30 minutes to make that more accurate.
    \item I believe that the first step of design should begin with free hand. The software tools are limit my creativity. That’s why mostly l was doing on paper sketch after that l import to revit or sketchup. I believe that style would be super useful.   \item I feel like if you get the rough shape + layout from your sketch, then you can easily import and get full 3D, vs. just sketching on paper and then you just have a flat sketch.
    \item (good to) start to visualize geometry in the plan into 3d prespective
    \item I don't tend to design cities/buildings, so this particular implementation likely wouldn't be useful for me personally, but a more general 3D-model-from-sketch (e.g. for random objects in a room, like a couch, table, etc.), could be useful for rapidly creating AR/VR spaces.
    \item Speed of making model and quick modify is useful
    \item This is too fun!
    \item (on sketching 2 views, and the possibility of sketching more) If I had to sketch a 2nd perspective, I wouldn't think it's worth it.
    \item If I had to sketch a 2nd perspective, I wouldn't think it's worth it.
    \item Nice to use. 
    \item The plan to perspective projection from perspective canvas to topdown is very useful when doing freefrom sketching.
\end{itemize}
}

\subsection{Quotes with negative sentiment} 
\emph{
\begin{itemize}
    \item The shape of the roof can (sic) be chosen. (likely meant can't)
    \item Only concern is how accurate it can be - I need details for only some situations
    \item It would be nice if I could edit the heights on the 3D model to what I wanted them to be (i.e. refine the 3D model by clicking and dragging on the tops of the buildings). I feel like it could also be useful to be able to quickly place trees and roads (things that aren't just buildings).
    \item Finer details are hard to sketch.
    \item Tried to draw pitched roof in the top-down: bad result. 
    \item I wish I could reduce the opacity of top-down projected sketch lines in the Perspective View. They're obscuring my reference image.
    \item Just like working in my sketchbook, but you also see the 3D even if it's not perfect
    \item Depending on the design scenario, I would want to sketch from different viewpoints (for the perspective sketch). For some scenarios, a street-level sketch would be better. But for massing, a higher perspective is better. But depending on the scene I am designing, I would like to change my sketch to match the scene: I wish I could change the viewpoint for perspective sketching in this tool.
\end{itemize}
}

\subsection{Quotes with neutral sentiment} 
\emph{
\begin{itemize}
    \item (Please add) Zoom in, zoom out tool
    \item (Please add) Line weight to differentiate elements in sketching
    \item To write text on it, e.g. overlay window, like a post-it note on the 3D mesh. Like annotation to show where the wind goes.
    \item keyboard shortcut to switch between canvases
    \item Maybe would want an image-to-sketch converter, so I can just pull in the image and then edit lines.
    \item Would be cool to also use sketches to define building details, e.g. door and window. Could be nice to use a prompt to texture the building.
    \item I like the melty thing it created - like gipsum - I couldn't do that in Rhino really. Rhino says: ``your line is this, follow it!''
    \item Details in the facade
    \item I think it would be nice to have a quick way to get a 3D representation to then get a more precise building. I could see myself tracing over with a cube [in the 3D view] - depends on the level of detail I'm going for. Normally in Blender, I'd start with a cube and position things relative to it. Could have a concrete wrapping of initial shape with a sharp convex hull. But could skip it if it's already sharp enough. 
    \item [UI] just needs small refinements
    \item If I want details, I'll just do it in Rhino.
\end{itemize}
}

\clearpage
\subsection{Summary of short-answer responses to the post-study questionnaire}
\label{sec-sup:post-study questionnaire}
\cref{tab:UserStudyPostQuestionsSummary} provides extended statistics supporting the discussion in \cref{sec:results_users} of the main paper.

\begin{table}[]
  \centering
  \resizebox{\columnwidth}{!}{%
      \begin{tabularx}{\textwidth}{|X|m{2cm}|m{2.5cm}|}
        \hline
        \textbf{Post-study Question} & \textbf{Architect cohort \newline (5 respondents)} & \textbf{non-Architect cohort \newline (5 respondents)} \\
        \hline
        Were you able to recreate the buildings in the reference image in 5 minutes? & Yes: 5/5 & Yes: 2/5 \newline No: 3/5 \\
        \hline
        How accurately were you able to recreate the buildings in the reference images using the sketching interface? 
Scale: [-2 -1 0 +1 +2] where +2: matches reference well & +1: 4/5 \newline 0: 1/5 & +2: 1/5 \newline +1: 4/5 \\
        \hline
        How likely are you to use the sketching interface in this study in the future for 3D building massing in the early design/ideation stage? Scale: [-2 -1 0 +1 +2], where 
-2: highly unlikely, +2: highly likely & +2: 3/5 \newline +1: 1/5 \newline  0: 1/5 & +1: 1/3 \newline 0: 2/3 \newline (2 non-responses) \\
        \hline
        Would you consider using the sketching interface in this study as part of your 3D model creation process? For example, instead of using 3D modeling software only (e.g. Rhino), ideating in this sketching interface, before  importing the output 3D mass building model into Rhino and continuing there? & Yes: 4/5 \newline No: 1/5 & Yes: 1/3 \newline Conditional Yes: 2/3 \newline (2 non-responses) \\
        \hline
        In the future, which one could you see yourself using for making 3D mass models? & Sketch: 3/5 \newline Rhino: 2/5 & Sketch: 3/5 \newline Rhino: 1/5 \newline Blender: 1/5 \\
        \hline
        Would seeing a 3D model from your sketch projected to  2D help you refine  your sketch? (overlaid in the 3D canvas) & Yes: 5/5 & Yes: 4/5 \newline No: 1/5 \\

        \hline
      \end{tabularx} }
      \caption{Summary of short-answer responses to the post-study questionnaire. Despite the sketch-based web interface being new for everyone, architects performed the task more swiftly on average. It is encouraging that three out of five architects were highly likely to use this sketching interface for massing, though non-architects were less enthusiastic.
      }
      \label{tab:UserStudyPostQuestionsSummary}
\end{table}

\section{User study: Comparison with SketchUp}
\begin{figure}[]
  \centering
  \includegraphics[width=0.98\linewidth]{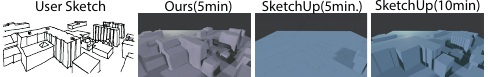}
   \caption{ One of four scenes modeled in GroundUp vs SketchUp by an architect. Qualitatively and quantitatively, quick progress is better in ours.}
   \label{fig:sketchUp}
\end{figure}

We tested two more architects, one of whom specializes in urban design. One uses SketchUp regularly; the other routinely works with similar software. 
Both were asked to model two scenes, first in our interface and then in SketchUp. 
In both systems, we provided top-down and perspective references. 
For SketchUp, we saved the output after 5 minutes and 10 minutes of modeling.
After 5 minutes in SketchUp, architects were able to only complete a flat outline of buildings. After 10 minutes, they were still not done with fixing the heights of the buildings, as shown in \cref{fig:sketchUp}. 
Meanwhile, with our system, architects were able to obtain 3D geometry in under 5 minutes. After massing, our models can be exported to detail-oriented modeling tools.

\section{Perspective depth prediction: Additional analysis and Visualizations}
\paragraph{Choice of $\nu$} 
In this section, we analyze the effect of different settings of $\nu$ values to construct occupancy feature volume. 

The 3D occupancy features are of shape $D \times H \times W$, where $D$ is the number of depth planes. 
When feeding these features into the 2D encoder in the UNet++, we consider depth planes as image feature channels $C$. We generate the occupancy features by setting all voxels that fall above non-occupied regions to $-\nu$ and all voxels above occupied regions to $\nu$.

We experiment with 5 different settings: $\nu \in \left\{ 1, 25, 50, 75, 100 \right\}$. 
\cref{table:ablation_v} shows that using $\nu=50$ performs better than the other settings. To understand the reason behind this, we observe the range of the multi-scale image features from the image encoder backbone. At the beginning of training, the range of these features is between 0 and 100 for the first few training batches. We believe that keeping $\nu$ close to the mid-point of that range allows the network to leverage the occupancy information most beneficially.

\begin{table}[]
\centering
\begin{tabular}{l|l|rrrrrr}
\multicolumn{1}{c|}{\textbf{}} & \multicolumn{1}{c|}{\textbf{Model}} & \multicolumn{1}{c}{\textbf{\begin{tabular}[c]{@{}c@{}}Abs \\ Diff↓\end{tabular}}} & \multicolumn{1}{c}{\textbf{\begin{tabular}[c]{@{}c@{}}Abs \\ Rel↓\end{tabular}}} & \multicolumn{1}{c}{\textbf{\begin{tabular}[c]{@{}c@{}}Sq \\ Rel↓\end{tabular}}} & \multicolumn{1}{c}{\textbf{RMSE↓}} & \multicolumn{1}{c}{\textbf{\begin{tabular}[c]{@{}c@{}}Log\\ RMSE↓\end{tabular}}} & \multicolumn{1}{c}{\textbf{a5↑}} \\ \hline
1 & $OV_{L-{\nu1}}$   & 4.64          & 3.39          & 0.31          & {\ul 7.04}    & 5.15          & 75.1          \\
2 & $OV_{L-{\nu25}}$  & 4.8           & 3.29          & 0.37          & 8.41          & 5.15          & 74.6          \\
3 & $OV_{L-{\nu50}}$  & \textbf{3.49} & \textbf{2.13} & \textbf{0.21} & \textbf{6.54} & \textbf{3.43} & \textbf{89.2} \\
4 & $OV_{L-{\nu75}}$  & 4.71          & 3.2           & 0.33          & 8.05          & 4.84          & 74.5          \\
5 & $OV_{L-{\nu100}}$ & {\ul 4.22}    & {\ul 2.68}    & {\ul 0.29}    & 7.74          & {\ul 4.2}     & {\ul 81.2}   
\end{tabular}%
\caption{The effect of the choice of $\nu$ for the Occupancy features Volume (OV) in the perspective depth prediction network. All models are trained using the ResNet-50 encoder. 
All metric values apart from $a5$ are scaled up by $10^2$.
}
\label{table:ablation_v}
\end{table}

\begin{figure}[]
  \resizebox{\columnwidth}{!}{\newcommand{\turnheightnew}{0.25\columnwidth}
\newcommand{\colone}{0000}
\newcommand{\coltwo}{0006}

\centering

\begin{tabular}{@{\hskip -1mm}c@{\hskip 1mm}c@{\hskip 1mm}c@{}}

\includegraphics[height=\turnheightnew]{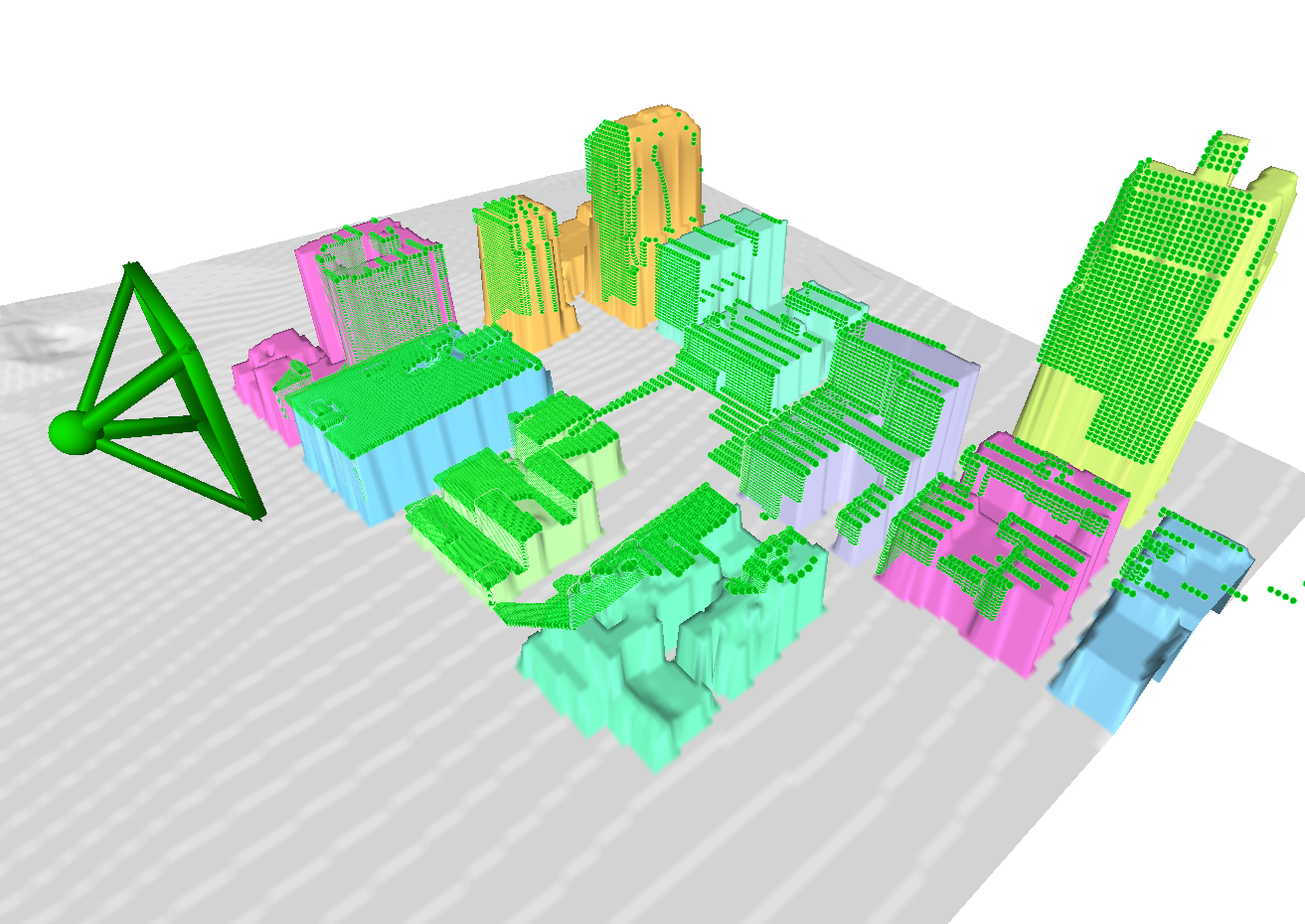} &
\includegraphics[height=\turnheightnew]{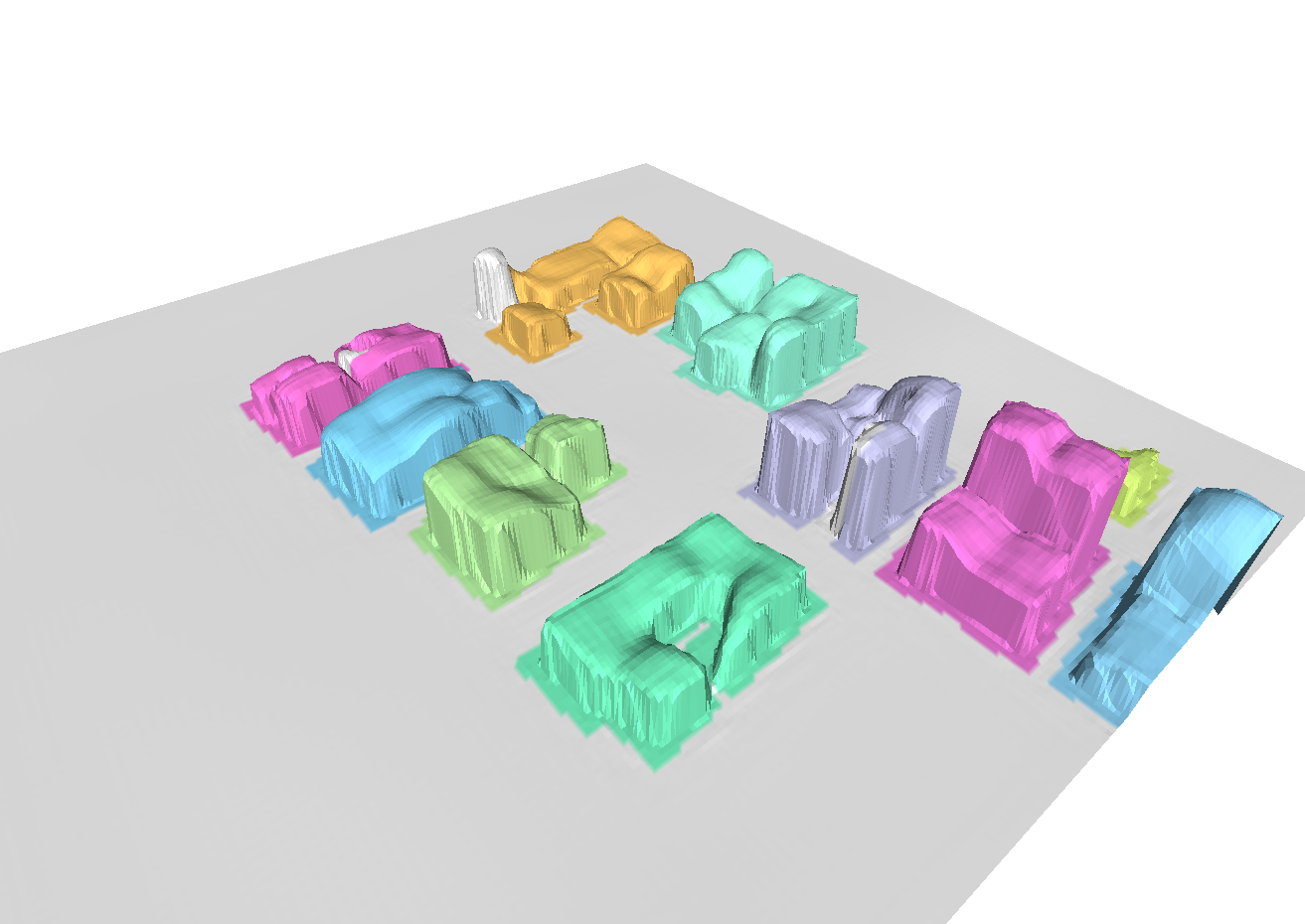} &
\includegraphics[height=\turnheightnew]{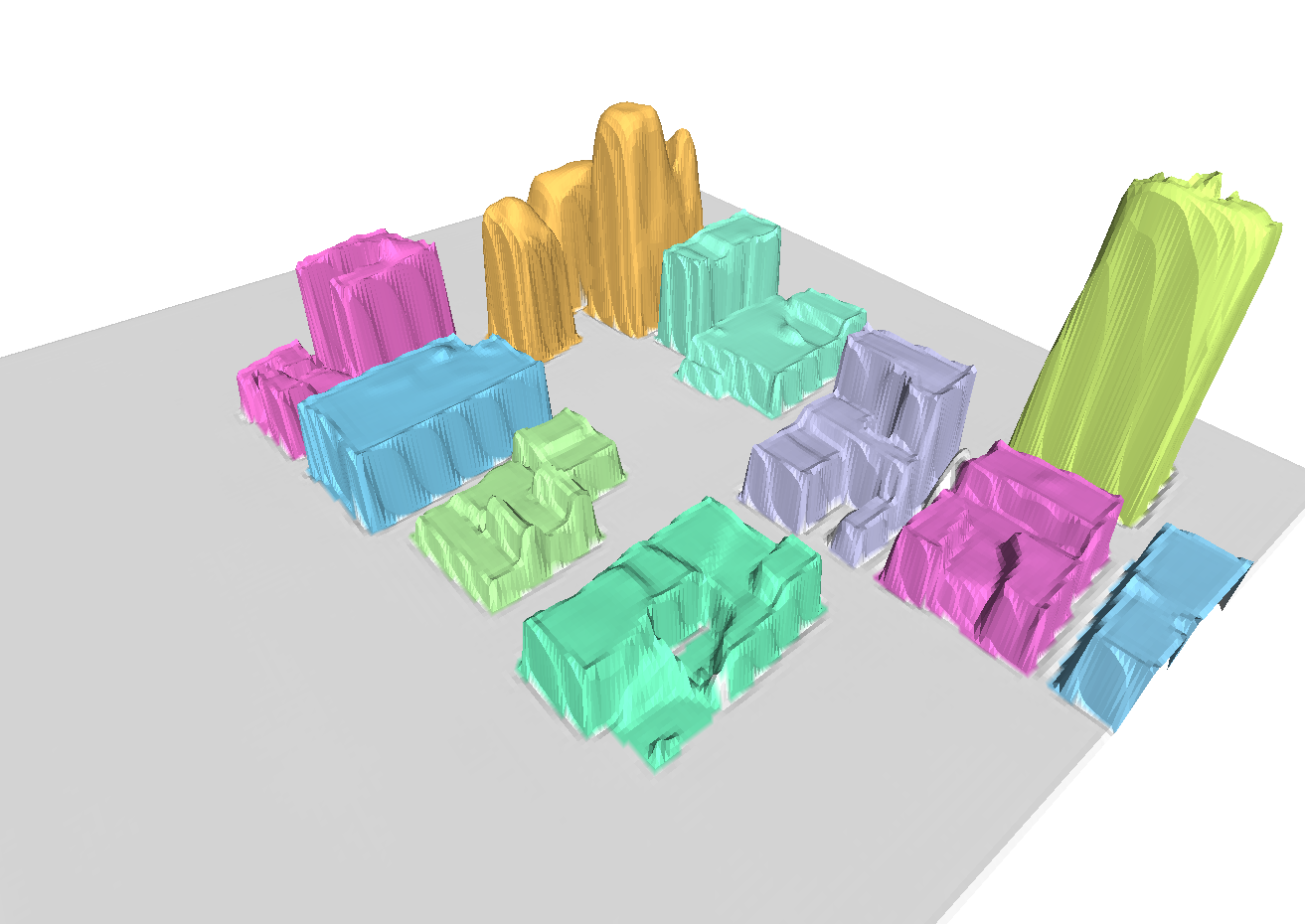} \\

a) & b) & c)  \\

\end{tabular}}
  \caption{Qualitative comparison of \textit{HeightFields}~\cite{watson2023wacv} vs our model. a) shows the ground-truth mesh with a camera marker for the perspective view and the visualization of what that view sees overlaid in green. b) is the \textit{HeightFields}~\cite{watson2023wacv} output and c) is our model.}
  \label{fig:ablation_diffusion}
\end{figure}

\paragraph{Sparse Height Information} \cref{fig:ablation_diffusion} shows height information our diffusion model gets as well as the baseline.

\section{Implementation details}

All our models and baselines were trained using PyTorch. For perspective depth prediction, we used a batch size of 16 across all models and ablation experiments, with a fixed learning rate of 1e-4 and weight decay. We trained all models for 25K iterations on
four RTX 2080 GPUs.
Our top-down mask model is trained with similar parameters to the depth predictor except we train it for 5K iterations. For building segmentation, we use Pytorch's \href{https://pytorch.org/docs/stable/generated/torch.nn.BCEWithLogitsLoss.html}{BCEWithLogitsLoss} function and set \textit{pos\_weight} as 20 for balancing the building pixels against the ground pixels in the mask image.
For our depth completion diffusion model, we set a batch size to $32$ and a learning rate to $3e-4$. 
We trained all models for $35$ epochs, on a machine with RTX 2080 GPUs.
For the depth completion baseline in the main paper, \textit{HeightFields}~\cite{watson2023wacv}, we used a batch size of $12$, a learning rate of $1e-4$, and trained it for 35 epochs on an NVIDIA RTX 3090 GPU. 
Please note that we added a normal loss $\mathcal{L}_\textrm{norm}$ to this baseline, as we found that results in more accurate reconstructions with sharper features. 
During training for our models and \textit{HeightFields}~\cite{watson2023wacv} baseline, we augment both the top-down and perspective sketches, following the strategy proposed by Ünlü \etal~\cite{gunlu3dvmannequin}. 

\subsection{Multi-conditional top-down diffusion model}
In the main paper, we describe how we condition the diffusion model in \cref{sec:design_diffusion}. 
Here, we provide additional details of the CNNs that we use to align features of the sketch and depth encoders.
The latent features  $c_\textrm{depth}$ and $z_{k}$ are passed through separate CNN heads: each head contains two convolutional blocks with a Conv2d layer followed by GroupNorm and Relu layers.

\section{Post-processing heightfields for visualization}

In the user interface, we use a quick (real-time) meshing algorithm.
We elevate each grid point of an initial 2D flat mesh using the predicted height values, as we described in \cref{sec:mesh_create}.

To obtain real-time performance, our predicted heightfields have limited spatial resolution, which results in some jagged aliased geometry on the vertical surface of the buildings. 
This effect can be observed for example in \cref{fig:ablation_diffusion}, in both the ground-truth and our predictions, which are both obtained from the same resolution heightfields.

To provide users with an option to work with higher-quality mesh at the next stage of their design pipeline, we explored automatic offline post-processing. 
We first vectorize the predicted heightfields using Adobe Illustrator's Image Trace tool.
We then export it as a high-resolution raster image (300dpi).
We use the following settings for Image Trace:
\begin{itemize}
    \item Preset: custom
    \item Mode: Grayscale
    \item Threshold: Between [8-20] (depending on the depth map, the threshold may vary.)
    \item Paths: 75 / Corners: 75 / Noise: 25 
\end{itemize}

For rendering the vectorized high-resolution output, we used Blender's Render Engine.
We used this approach to generate visualizations in the teaser in the main paper (\cref{fig:teaser}), the supplementary video, and for the visualization of the results of the freehand modeling sessions (\cref{fig:user_study_reference} in the main paper).

Potentially, some superresolution approaches that do not require training, such as \cite{du2023demofusion}, can also be used to reduce the jaggedness of the reconstructed meshes.

\section{\new{Controllable geometry generation in occluded areas}}
As sketching is typically the first step in any design process, our primary goal was to enable a tool that combines the benefits of sketching and 3D shape exploration for large-scale city scenes. Depending on the use case, the user interface could be modified to fit different modeling scenarios.
Our UI could be extended to allow modeling buildings 1-1 -- the strategy chosen during sketching by one of our participants in the user study, \textit{Novice-User-2}. 
For another scenario, the UI could evolve to support multi-view perspective sketches. 
Furthermore, one user asked for camera-angle control (see \cref{sec-sup:post-study questionnaire}).

While we leave a thorough exploration of multi-view iterative editing to future work, we have conducted a preliminary study.
To test this, we used 250 test scenes with 2 perspective views $45^\circ$ apart. 
We projected point clouds inferred from extra perspective sketches into the top-down representation passed to our diffusion model. 
Without any finetuning, the reconstruction is improved on all metrics, \eg by $.0020$ points on absolute difference of top-down view, compared to a single perspective sketch, as seen in \cref{table:diff_metrics_multi}. 

\section{\new{Effect of normals loss}}

\begin{table}[]
\centering
\begin{tabular}{lccccc}
\cline{1-5}
\textbf{Model} & \textbf{Abs Diff↓} & \textbf{Abs Rel↓} & \textbf{Sq Rel↓} & \textbf{RMSE↓} \\ \hline
GroundUp (single camera)& 0.1158 & 0.0249 & 0.0073 & 0.1619 \\
GroundUp (two cameras) & \textbf{0.1138} & \textbf{0.0244} & \textbf{0.0069} & \textbf{0.1557} \\ \hline
\cline{1-5}
\end{tabular}
\caption{Quantitative evaluation on multi-view input. GroundUp with multi-view input improves metrics.}
\label{table:diff_metrics_multi}
\end{table}

In \cref{table:diff_3D_metrics} of the main paper, we noticed a drop in performance when the normal loss is used. 
We tracked this down through visualizations - please see \cref{fig:normal-loss-m}. 
This loss causes geometry to shrink slightly in all directions - especially in the areas occluded in the perspective sketch. 
In \cref{fig:normal-loss-m}-a, the red point cloud accounts for both visibility and actual building height. 
For $\mathcal{L}_{t,norm}$, \cref{fig:normal-loss-m}-b shows the red point cloud is riding slightly above the green prediction, meaning the height is underestimated. 
In contrast, the prediction without the normal loss does not suffer from underestimated heights within the visibility region, albeit producing uneven surfaces (\cref{fig:normal-loss-m}-c). $\mathcal{L}_{t,norm}$ produces nice building geometry with even surfaces within and outside the visibility region; without it, the model produces unrealistic buildings, deviating a lot from real building geometry, especially outside the visibility region (\cref{fig:normal-loss-m}-c).  

The 3D metric in \cref{table:diff_3D_metrics} masks for visibility, so this metric is sensitive to shrinkage while ignoring defects outside the perspective view. The 2D metrics are computed for the full buildings' geometries and reflect on the quality of buildings' rooftops outside of areas visible in the perspective views. 

We think the reason for the \old{performance drop }\new{geometry shrinkage in the visible regions }could be explained with the aid of multi-task learning literature. Training a neural network with an auxiliary task could affect the performance on the main task, \eg that of depth and normals estimation \cite{fifty2021efficiently}.

\end{document}